\title{The Word Entropy of Natural Languages}
\author{
        Christian Bentz \\
        Department of Linguistics\\
        University of T\"{u}bingen\\        
        \vspace{0.5cm} 
        T\"{u}bingen 72074, Germany\\          
        Dimitrios Alikaniotis\\
        Department of Theoretical and Applied Linguistics\\
        University of Cambridge\\
        Cambridge, UK\\        
}

\date{\today}

\documentclass[12pt]{article}
\usepackage{latexsym}
\usepackage{graphicx}
\usepackage{amssymb}
\usepackage{color,soul}
\usepackage{amsmath}
\usepackage[utf8]{inputenc}
\usepackage{fixltx2e}
\usepackage{pdflscape}

\DeclareRobustCommand{\hlcyan}[1]{{\sethlcolor{cyan}\hl{#1}}}

\begin{document}
\maketitle

\begin{abstract}
The average uncertainty associated with words is an information-theoretic concept at the heart of quantitative and computational linguistics. The entropy has been established as a measure of this average uncertainty - also called average information content. We here use parallel texts of 21 languages to establish the number of tokens at which word entropies converge to stable values. These convergence points are then used to select texts from a massively parallel corpus, and to estimate word entropies across more than 1000 languages. Our results help to establish quantitative language comparisons, to  understand the performance of multilingual translation systems, and to normalize semantic similarity measures.
\end{abstract}

\section{Introduction}
The predictability of symbols in strings is the most fundamental concept of information encoding in general, and natural language production in particular. Shannon \cite{Shannon1949} defined the \textit{entropy} - or \textit{average information content} as a measure of uncertainty or ``choice'' inherent to strings of symbols. Since then, Shannon \cite{Shannon1951} and others have undertaken great efforts to estimate precisely the entropy of written English \cite{Brown1992,Kontoyiannis1998,Gao2008,Schurmann1996}, and other languages \cite{Behr2003,Montemurro2011,Montemurro2015}. 

In computational linguistics, entropy - and related measures - have been widely applied to tackle problems relating to machine translation \cite{Berger1996,Och2002}, distributional semantics \cite{Herbelot2013,Pado2015,Santus2014,Resnik1995}, information retrieval \cite{Boon2014,Stetson2002,McFarlane2009}, and multiword expressions \cite{Zhang2006,Ramisch2008}. All these accounts crucially hinge upon estimating the probability and uncertainty associated with words - i.e. the \textit{word entropy} - in a given text and language.  

There are two central questions associated with this estimation: 1) what is the text size (in number of tokens) at which word entropies converge to a stable value? 2) How much systematic difference in word entropies do we find across different languages? The first question is related to the problem of data sparsity. The performance of NLP tools relying on word probabilities is lower-bounded by the minimum text size at which word entropies can still be reliably estimated. The second question relates to the applicability of NLP tools across different languages. The performance of a single tool can be systematically biased in a specific language due to generally higher or lower word entropies. 

In this study, we use a state-of-the-art method to estimate block entropies \cite{Nemenman2001}, and also implement a source entropy estimator \cite{Gao2008}. This allows us to establish word entropy convergence points for parallel texts of up to 30M tokens in 21 languages. Based on these analyses, we select texts with sufficiently large token counts from a massively parallel corpus and estimate word entropies across 1360 texts and 1001 languages. 

\section{Motivation}\label{motivation}
A fundamental property of words is their frequency of occurrence, and hence their probability in language production. This probability is at the heart of many applications in natural language processing. 

For example, the probability of an expert translating the English word \textit{the} into German \textit{der} can be estimated as

\begin{equation}\label{eq:prob}
\hat{p}(der|the)= \hat{p}(der,the)/\hat{p}(the).\\ 
\end{equation} 
This is the conditional probability of finding $der$ in a word aligned German translation where we have $the$ in the English original. More generally, we could have a set of possible German translations 

\begin{equation}
\mathcal{S}=\{der,die,das,dem,den,des\}
\end{equation}

assigned with estimated probabilities: 

\begin{equation}
\hat{p}(der) + \hat{p}(die) + \hat{p}(das) + \hat{p}(dem) + \hat{p}(den) + \hat{p}(des) = 1.
\end{equation}

In the uniform case, all of these are assigned probability 1/6, and there is maximum uncertainty about which German word corresponds to English \textit{the}. In the clearest case, one of the probabilities is 1 and the others are 0. The entropy over such a distribution of word probabilities measures exactly this uncertainty or ``choice''. In our example of German articles, the word entropy is $\approx2.6$ in the uniform case and 0 in the clear case. Entropy calculation is detailed below (Section 4). 

\subsection{Entropy in statistical machine translation}
Given the example above, the task of a machine translation system is to map the English word $the$ onto the correct choice from the set of German translations. The maximum entropy approach \cite{Berger1996}, for instance, requires that the probabilities are estimated as accurately as possible from the given training data. Probabilities which cannot be estimated reliably are then chosen in a way to maximize the overall entropy of the model. 

Crucially, the translation difficulty in such an account is a direct function of the entropy of the word distribution. In the English-to-German mapping - with uniform probabilities - there are 2.6 bits of ``choice'', whereas in the German-to-English mapping there are 0 bits of ``choice''. 

This is closely related to Koehn's \cite{Koehn2005} finding that when translating into English from 10 different source languages, BLEU scores negatively correlate with the number of words in the source language. Hence, it is a harder problem to translate into a language with higher word entropies than to translate into a language with lower word entropies - everything else being equal. From this perspective, estimating the entropy of word distributions can help to predict translation performance.

\subsection{Entropy in distributional semantics}
Measures of semantic content and similarity \cite{Resnik1995,Santus2014,Pado2015,Herbelot2013} often also rely on information-theoretic concepts. For example, the classical study by Resnik \cite{Resnik1995} defines the similarity between two concepts $c_1$ and $c_2$ in a semantic hierarchy as

\begin{equation}\label{eq:semsim}
sim(c_1,c_2)= \max_{c\;\in\;S(c_1,c_2)} [-log\;p(c)], \\ 
\end{equation} 
with $S(c_1,c_2)$ being the set of concepts which subsume both $c_1$ and $c_2$. To get the information content $-log\;p(c)$, the probability of a shared concept $p(c)$ is  estimated from corpus data as 

\begin{equation}\label{eq:freq}
\hat{p}(c)=freq(c)/N, \\ 
\end{equation} 
where $freq(c)$ is the count of word tokens (nouns in this case) per concept (e.g. occurrences of \textit{dollar}, \textit{euro}, and \textit{coin} would count towards the frequency of the concept \textit{money}), and $N$ is the overall number of word tokens observed. Note that via the estimated $\hat{p}(c)$ this similarity measure depends on the overall distribution of word probabilities, i.e. the word entropy. Namely, for a language with more word types of overall lower token frequencies the probability $\hat{p}(c)$ is biased to be lower on average, and hence the information content $-log\;p(c)$ is biased to be higher. 

Similar considerations apply to finding hypernyms in vector spaces \cite{Santus2014}, measuring semantic content to establish asymmetries between derived words and their base forms \cite{Pado2015}, as well as measuring differences in semantic content to establish hyponym-hypernym relations \cite{Herbelot2013}.  

Overall, estimating a) the convergence points (in number of word tokens), and b) reliable approximations for word entropies are crucial prerequisites for understanding the performance of translation systems, semantic similarity measures, and more generally, any NLP tool that relies on probabilities of words. 

These estimations are also relevant for efforts to broaden the scope of NLP to lesser known languages \cite{Lewis2010,Oestling2014,Petrov2011,Marneffe2014}, and to establish quantitative and corpus-based methods in linguistic typology \cite{Cysouw2007,Waelchli2012a,Waelchli2012b}.   

\section{Data}
To control for constant content across languages, we use two sets of parallel texts: 1) the \textit{European Parliament Corpus} (EPC) \cite{Koehn2005}, and 2) the \textit{Parallel Bible Corpus} (PBC) \cite{Mayer2014}.\footnote{Last accessed on 09/03/2016} Details about the corpora can be seen in Table~\ref{pcorpora}. The general advantage of the EPC is that it is big in terms of numbers of word tokens per language (ca. 30M), whereas the PBC is smaller (ca. 280K word tokens per language), but massively parallel in terms of encompassing $>1000$ languages. 

\begin{table}
\small
\centering
\begin{tabular}{|l|llll|}
\hline \bf Parallel Corpus & \bf Size & \bf $\varnothing$ Size & \bf Texts & \bf Lang. \\ \hline
EPC & $\approx600M$ & $\approx30M$ & 21 & 21 \\
PBC & $\approx420M$ & $\approx280K$ &  1471 & 1083 \\
\hline
\end{tabular}
\caption{\label{pcorpora} Information on the parallel corpora used.}
\end{table}

\section{Methods}
The basic information encoding unit chosen in this study is the word. Earlier studies on the entropy of English \cite{Shannon1951,Kontoyiannis1998,Schurmann1996} often chose letters instead. However, in computational linguistics, word tokens are are a common working unit.

A \textit{word token} is here defined as a string of alphanumeric UTF-8 characters delimited by white spaces, with all letters converted to lower case and punctuation removed. Note that scripts of Mandarin Chinese (cmn) and Khmer (khm), for instance, delimit phrases and sentences by white spaces, rather than words. However, such scripts constitute a negligible proportion of our sample ($\approx0.01$\%). In fact,  $\approx90$\% of the texts are written in Latin script.

Given words as basic information encoding units, we can estimate the word entropy as outlined in the following.   

\subsection{Entropy estimation}
Assume a text is a random variable $T$ created by a process of drawing and concatenating tokens from a set (or vocabulary) of word types $\mathcal{V}=\{w_1,w_2,...,w_V\}$, with vocabulary size $V=|\mathcal{V}|$. Word type probabilities are distributed according to $p(w)=Pr(T=w)$ for $w \in \mathcal{V}$. Given these definitions, the entropy of $T$ can be calculated as \cite{Shannon1949}

\begin{equation}\label{eq:entropy}
H(T) = - \sum_{i=1}^V p(w_i) \log_2  ( p(w_i ) ). \tag{4} 
\end{equation}    
$H(T)$ can be seen as the \textit{average information content} of word types. A crucial step towards estimating $H(T)$ is to reliably approximate the probabilities $p(w_i)$. 

\subsection{Block entropies}
In a text, each word type $w_i$ has a token frequency $f_i=freq(w_i)$.
Take the first verse of the English Bible as a text. 
\\
\\
{\fontfamily{qcr}\selectfont
in \hl{the} beginning god created \hl{the} heavens \hlcyan{and} \hl{the} earth\\
\hlcyan{and} \hl{the} earth was waste \hlcyan{and} empty [...]}
\\
\\
In this example, the word type \textit{the} occurs 4 times, \textit{and} occurs 3 times, etc. As a simple approximation, $p(w_i)$ can be estimated via the maximum likelihood method:

\begin{equation}
\hat{p}(w_i) = \frac{f_i}{\sum_{j=1}^V f_j}, \tag{5} 
\end{equation}    
where the denominator is the overall number of word tokens.
For the Bible verse we would thus have:

\begin{equation}
H(T) = - ( \frac{4}{17} \log_2 ( \frac{4}{17} ) + \frac{3}{17} \log_2 ( \frac{3}{17} ) + \dots \\ + \frac{1}{17} \log_2 ( \frac{1}{17} ) ) \approx 3.2 \tag{6}
\end{equation}    
However, there are two main caveats with this so-called \textit{plug-in} approach: 

Firstly, it has been shown that the maximum likelihood estimator is unreliable, especially for small text sizes \cite{Nemenman2001,Hausser2009}, i.e. small $\sum_{j=1}^{V} f_j$. A range of remedies have been proposed to overcome this problem \cite{Schurmann1996,Nemenman2001,Hausser2009,Lesne2009}. 

Secondly, estimating the entropy from raw counts assumes that word tokens are drawn independently from a multinomial distribution, meaning there are no dependencies between words. Clearly, this requirement is not met for natural languages \cite{Lesne2009}. 

To overcome this problem, instead of using unigrams as ``blocks'' of information encoding, we could use bigrams, trigrams, n-grams, and thus increase block sizes to 2, 3, n. This yields \textit{block entropies} \cite{Schurmann1996} defined as

\begin{equation}\label{eq:blockentropy}
H_n(T)=-\displaystyle \sum_{i=1}^{V} p(w_i,w_{i+1},...,w_n) \times \\ \log_{2}(p(w_i,w_{i+1},...,w_n)),
\end{equation}  
where $n$ is the block size. If $n$ is big enough to take into account most long-range correlations between word tokens, then $H_n(T)$ is a close approximation of $H(T)$. However, since the number of different blocks grows exponentially with $n$, very big corpora are needed to get reliable estimates. Note that Schürmann \& Grassberger \cite{Schurmann1996} use an English corpus of 70M words and assert that entropy estimation beyond a block size of 5 letters (not words) is already unreliable. We will therefore stick with block sizes of 1, i.e. \textit{unigram entropies} here.

However, we implement a more parsimonious approach taking into account dependencies between words. This is based on the theory behind Lempel-Ziv compression \cite{Ziv1977,Ziv1978}. 

\subsection{Source entropies}
Instead of calculating $H_n(T)$ with ever increasing block sizes $n$, Kontoyiannis et al. \cite{Kontoyiannis1998} and Gao et al. \cite{Gao2008} suggest to use the findings on optimal compression by Ziv \& Lempel \cite{Ziv1977,Ziv1978}. 

More precisely, \cite{Gao2008} show that entropy estimation based on the so-called \textit{increasing window estimator}, or \textit{LZ78 estimator} \cite{Cover2006}, is efficient in terms of convergence. 

Applied to the problem of estimating word entropies, the method works as follows: for any given word token $t_i$ in a text find the longest match-length $l$ for which the string $s=(t_i,t_{i+1},...,t_{i+l})$ matches a preceding string in $(t_1,...,t_{i-1})$. Formally, define $l$ as 

\begin{equation}\label{eq:matchLength}
l_i=1+\max\{0\leq l \leq i: s_{i}^{i+l-1}=s_{j}^{j+l-1} \\ for \; some \; 0 \leq j \leq i-1\}.
\end{equation}  
This is an adaptation of Gao et al.'s \cite{Gao2008} match-length definition.\footnote{Note that Gao et al. \cite{Gao2008} give a more general definition that also holds for the so-called \textit{sliding window}, or \textit{LZ77} estimator.}
To illustrate this, take the example of the English Bible again:
\\
\\
{\fontfamily{qcr}\selectfont
in\textsubscript{$1$} the\textsubscript{$2$} \hl{beginning}\textsubscript{$3$} god\textsubscript{$4$} created\textsubscript{$5$} the\textsubscript{$6$} heavens\textsubscript{$7$} \underline{and}\textsubscript{$8$} \underline{the}\textsubscript{$9$} \\ \underline{earth}\textsubscript{${10}$} \hl{and}\textsubscript{${11}$}
\underline{the}\textsubscript{${12}$} \underline{earth}\textsubscript{${13}$} was\textsubscript{${14}$} waste\textsubscript{${15}$} and\textsubscript{${16}$} empty\textsubscript{${17}$} [...]}
\\
\\
For the word token \textit{beginning}, in position $3$, there is no match in the preceding string (\textit{in the}). Hence, the match-length $l_3$ is $0(+1)=1$. If we look at \textit{and} in position $11$, then the longest matching string is \textit{and the earth}. Hence, the match-length $l_{11}$ is $3(+1)=4$.  

Note that the average match-lengths across all word tokens reflect the redundancy in the string - which is the inverse of unpredictability. Based on this connection, Gao et al. \cite{Gao2008} (Equation 6) show that the entropy of the string can be approximated as 

\begin{equation}\label{eq:sourceEntropy}
\tilde{H}(T)=\frac{1}{N} \displaystyle \sum_{i=2}^{N} \frac{\log_2(i)}{l_i}, 
\end{equation}   
where $N$ is the overall number of tokens, and $i$ is the position in the string. This approximates the \textit{entropy rate}, or \textit{per-symbol entropy} \cite{Gao2008}, which is denoted as $h$ by Lesne et al. \cite{Lesne2009}, and for which holds

\begin{equation}\label{eq:h}
h=\lim_{N \to \infty} H(t_{N+1}|t_1,t_2,...,t_N).
\end{equation} 
In other words, as the number of tokens $N$ approaches infinity, $h$ reflects the average information content of a token $t_{N+1}$ conditioned on all preceding tokens. So $h$ accounts for \textit{all statistical dependencies} between tokens \cite{Lesne2009}. 

We will call $h$ and its approximation $\tilde{H}(T)$ the \textit{source entropy} - after Shannon's \cite{Shannon1949} formulation of \textit{the entropy of an information source}.\footnote{Note that in the limit, i.e. as block sizes $n$ approach infinity, \textit{block} and \textit{source} entropies are the same. Also, for an independent and identically distributed (i.i.d) random variable, the block entropy of block size 1, i.e. $H_1(T)$, is identical to the source entropy $h$ \cite{Lesne2009}. However, as pointed out above, in natural languages words are not independently distributed.}  

\begin{landscape}
\begin{figure*}[htbp!]
\centering \includegraphics[width=1.5\textwidth]{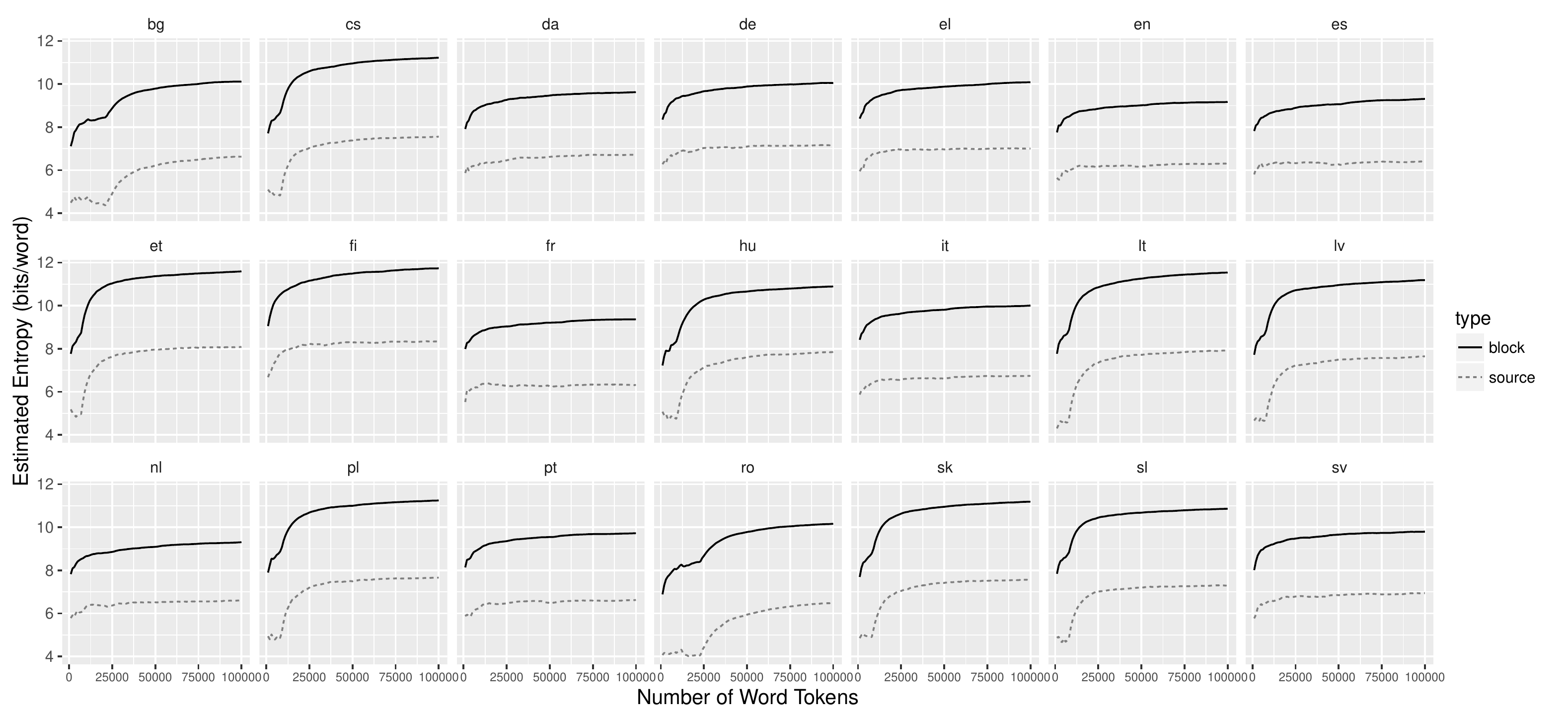}
\caption[]{\label{plot:Hconv} Entropy convergence across 21 languages of the EPC corpus. The number of tokens on the x-axis is limited to 100K, since entropy values already converge before that. Language identifiers are given above the panels: bg: Bulgarian, cs: Czech, da: Danish, de: German, el: Modern Greek, en: English, es: Spanish, et: Estonian, fi: Finish, fr: French, hu: Hungarian, it: Italian, lt: Lithuanian, lv: Latvian, nl: Dutch, pl: Polish, pt: Portuguese, ro: Romanian, sk: Slovak, sl: Slovene, sv: Swedish.}
\end{figure*}
\end{landscape}

\begin{landscape}
\begin{figure*}[htbp!] 
\centering \includegraphics[width=1.5\textwidth]{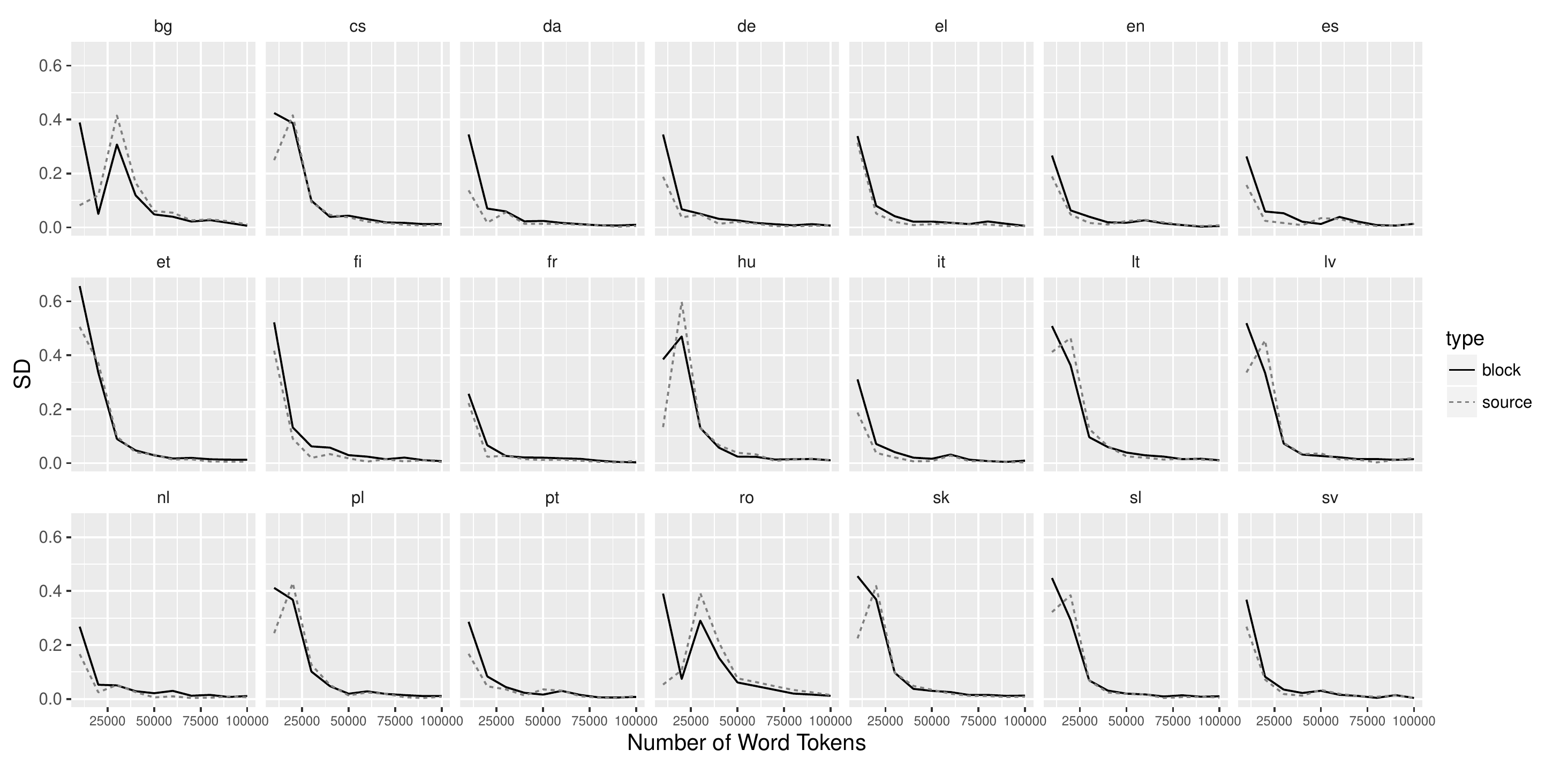}
\caption[]{\label{plot:SDconv} Convergence of SDs of entropies (y-axis) across 21 languages of the EPC corpus. Average SD values are calculated at each 10K tokens.}
\label{}
\end{figure*}
\end{landscape}

\subsection{Implementation}
We estimate \textit{unigram entropies} - i.e. entropies for block sizes of 1 ($H_1(T)$) - using the Python implementation of the \textit{Nemenman-Shafee-Bialek} (NSB) \cite{Nemenman2001} estimator. This estimator has a faster convergence rate compared to other block entropy estimators \cite{Hausser2009}.\footnote{https://gist.github.com/shhong/1021654/} 

Moreover, we implement the source entropy estimator $\tilde{H}(T)$ as in Gao et al.'s \cite{Gao2008} proposal. This is inspired by an earlier implementation by Montemurro \& Zanette \cite{Montemurro2011} of Kontoyiannis et al.'s \cite{Kontoyiannis1998} estimator. The code of this implementation will be made available on \textit{github}. 

\section{Results}
We first assess the text sizes at which both block and source entropies converge to a stable value using a subset of 21 languages (Section~\ref{Hconv}). We then select texts from the full PBC corpus based on the minimum number of tokens needed for convergence, estimate their entropies, and compare their spread on an entropy spectrum (Section~\ref{Hest}). Finally, in Section~\ref{Hcor}, we investigate the correlation between block and source entropies.   

\subsection{Entropy convergence}\label{Hconv}
Figure~\ref{plot:Hconv} illustrates the convergence of block and source entropies across 21 languages of the EPC. Note that block entropies are generally higher than source entropies. This is expected given that uncertainty is reduced for source entropies by taking the preceding co-text into account. 

We further establish convergence points by calculating SDs of entropies for step sizes of 10K tokens, illustrated in Figure~\ref{plot:SDconv}. If we choose $SD<0.05$ as a convergence criterion, then we get the convergence points per language given in Table~\ref{table:Hconv}. 

\begin{table}
\small
\centering
\begin{tabular}{|l|ll|}
\hline \bf Language & \bf Source $H$ & \bf Block $H$ \\ \hline
bg & 70000 & 20000 \\
cs & 40000 & 40000 \\
da & 20000 & 40000 \\
de & 20000 & 40000 \\
el & 30000 & 30000 \\
en & 20000 & 30000 \\
es & 20000 & 40000 \\
et & 40000 & 40000 \\
fi & 30000 & 50000 \\
fr & 20000 & 30000 \\
hu & 50000 & 50000 \\
it & 20000 & 30000 \\
lt & 50000 & 50000 \\
lv & 40000 & 40000 \\
nl & 20000 & 40000 \\
pl & 50000 & 40000 \\
pt & 20000 & 30000 \\
ro & 70000 & 60000 \\
sk & 40000 & 40000 \\
sl & 40000 & 40000 \\
sv & 30000 & 30000 \\
\hline
$\varnothing$ & $\approx\textbf{35K}$ & $\approx\textbf{38K}$ \\
\hline
\end{tabular}
\caption{\label{table:Hconv} Convergence points ($SD<0.05$) in number of tokens for source and block entropies.}
\end{table} 

Note that all 21 languages converge to stable entropies below text sizes of 100K tokens, with a maximum of 70K tokens (bg and ro) and an average of 35K for source and 38K for block entropies. This is an encouraging result, since texts with a minimum of around 70K tokens are available for a wide range of languages in the PBC.   

\subsection{Entropy estimates}\label{Hest}
Based on the convergence analyses, we choose 100K tokens as a cut-off point for inclusion of PBC texts. This leaves us with 1352 texts for block entropies, and 1360 texts for source entropies.\footnote{The number of texts is lower for block entropies since estimations failed for 8 texts due to some punctuation marks that were not correctly removed.} Both cover 1001 languages. Figure~\ref{H_density} is a density plot of the estimated entropy values. Block entropies are approximately normally distributed around a mean of $9.26$ ($SD=1.24$), and source entropies around a mean of $5.97$ ($SD=1.07$). Again, source entropies are systematically lower than block entropies.    

\begin{figure}[htbp!] 
\centering \includegraphics[width=0.8\textwidth]{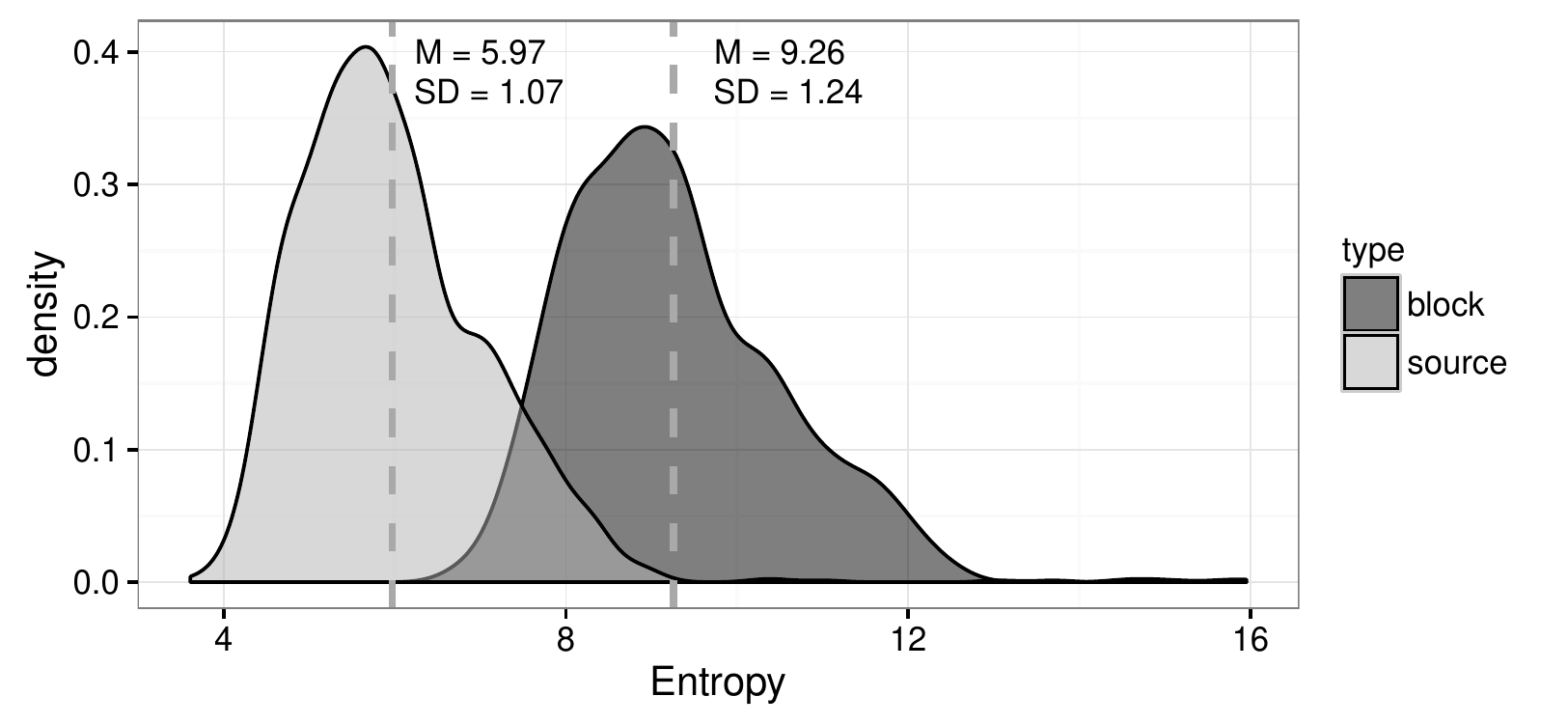}
\caption[]{\label{H_density} Density plot of block and source entropies across texts of the PBC with $>100K$ tokens, amounting to 1360/1352 texts and 1001 languages.}
\label{H_density}
\end{figure}

It is remarkable that given the wide range of potential entropies - from 0 to $>$16 - most natural languages fall on a relatively narrow spectrum. For example, block entropies mainly fall in the range between 7 and 12, thus only covering around 30\% of the possible range.  
 
\subsection{Correlation between block and source entropies}\label{Hcor}
\begin{figure}[htbp!] 
\centering \includegraphics[width=0.8\textwidth]{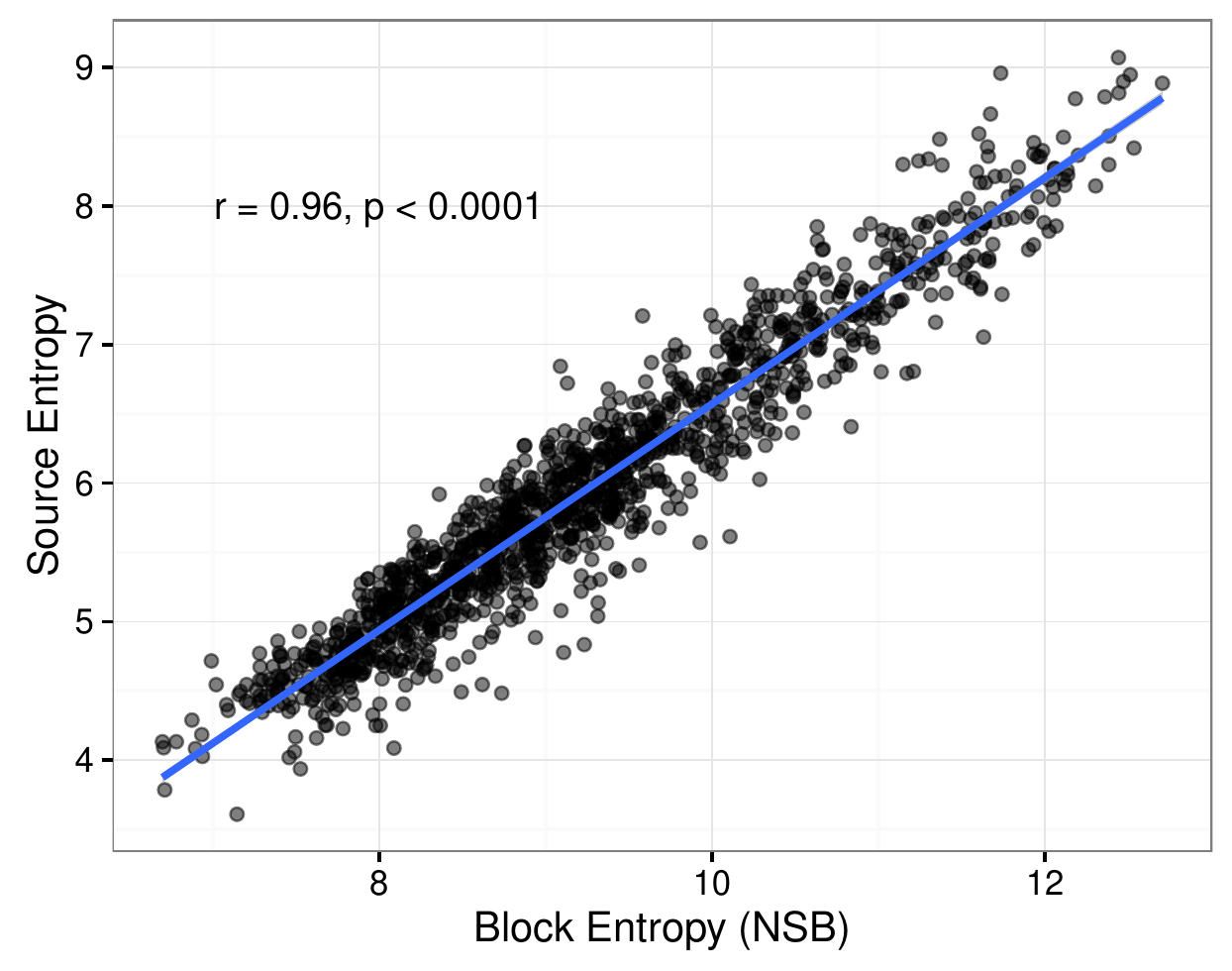}
\caption[]{\label{plot:Hcor} Correlation between block and source entropies for the PBC texts.}
\end{figure}

The similarities in convergence lines in Figure~\ref{plot:Hconv} suggest that there is a correlation between block and source entropy estimates.  Figure~\ref{plot:Hcor} elicits this correlation by plotting block entropies (x-axis) versus source entropies (y-axis). The Pearson correlation is strong ($r=0.96, p<0.0001$), suggesting that despite the differences in the estimation methods there is a strong connection between them.\footnote{Five clear outliers were removed here. These are texts of languages like Chinese (cmn) and Khmer (khm), which rather delimit phrases and sentences by white spaces. These have incommensurable source and block entropies.}   

\section{Discussion}
Independent of the estimation method - using block or source entropies - texts of 100K tokens are generally sufficient to estimate values reliably. This was shown across 21 languages of the EPC. Of course, this is not to say that there are no texts/languages in a bigger corpus like the PBC for which convergence might take longer. Note that the 21 EPC languages cover around $50$\% of the full range of values across the PBC sample. For example, block entropies range from $\approx 9$ for English to $\approx 12$ for Finnish in the EPC, and from $\approx 6$ to $\approx 13$ in the PBC. 

Moreover, there is a strong correlation between block and source entropies. This is somewhat surprising considering that the probabilities of word occurrences are generally assumed to be strongly dependent on co-text.  
Of course, there is a co-text effect. It yields source entropies which are lower than block entropies. However, the difference between them is systematic and allows us to predict source entropies from unigram block entropies. Namely, a linear model fitted through the points in Figure~\ref{plot:Hcor} can be specified as

\begin{equation}\label{eq:convertH}
\tilde{H}(T)=-1.59+0.82\, H_1(T). 
\end{equation} 
Via Equation~\ref{eq:convertH} we can convert block entropies into source entropies with a mean difference of $0.03$. Note that estimating source entropies requires searching strings of length $i-1$. As $i$ increases, the CPU time per additional word token increases linearly, whereas unigram block entropies can be estimated based on dictionaries of word types and their token frequencies, and the processing time per additional word token is constant. Hence, Equation~\ref{eq:convertH} can help to reduce processing cost.

\subsection{Predicting translation performance}
Based on estimated entropies per language, we can predict the difficulty of pairwise translations between languages, i.e. translation system performance. 

\begin{figure}[htbp!] 
\centering \includegraphics[width=0.8\textwidth]{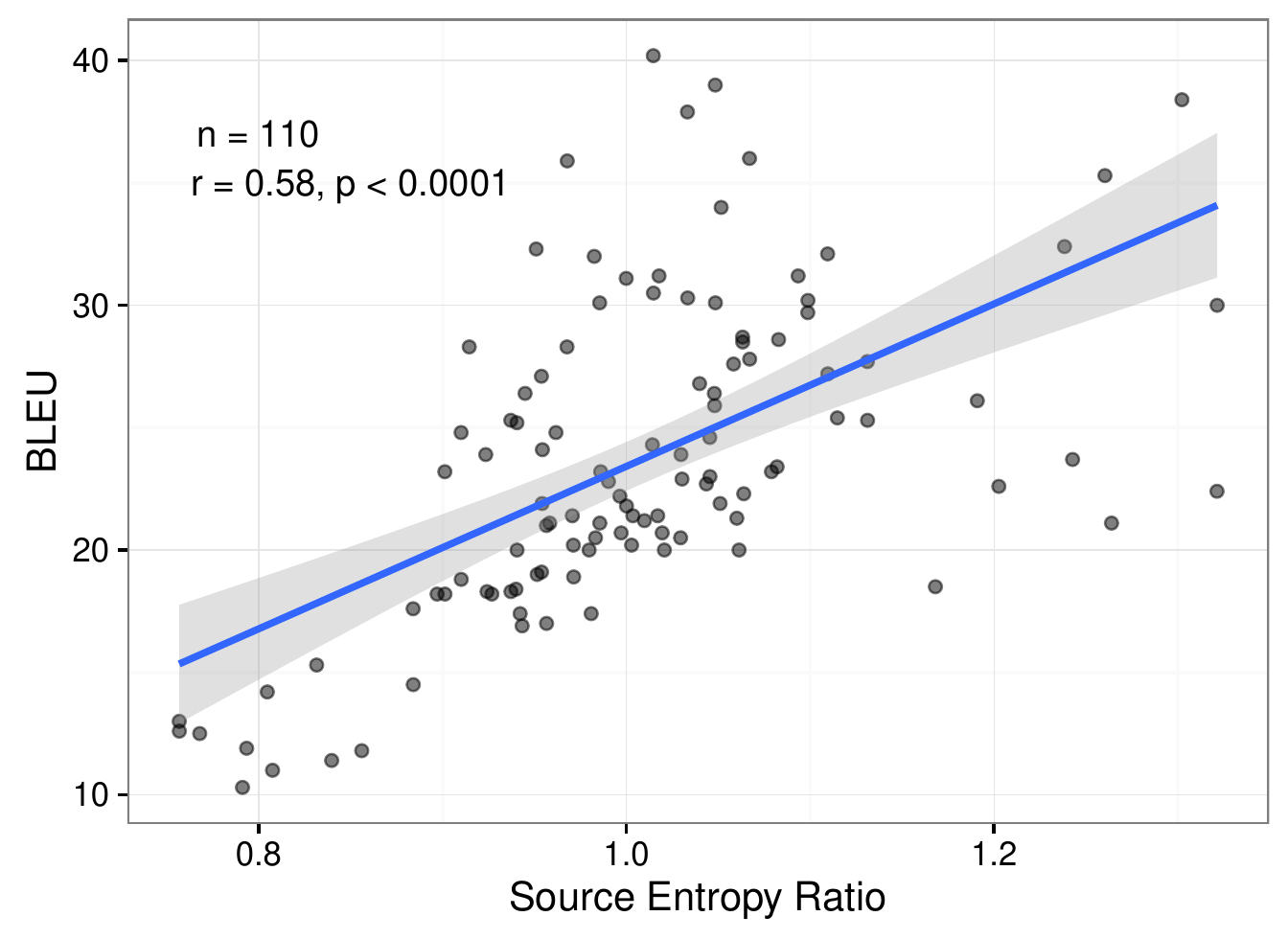}
\caption[]{\label{plot:BLEUcor} Correlation between ratios of source entropies and BLEU scores of a statistical machine translation system \cite{Koehn2005}. Entropy ratios correspond to pairwise ratios for 11 languages of the PBC:\textit{ da, de, el, en, es, fi, fr, it, nl, pt, sv}.}
\end{figure}

\cite{Koehn2005} used a probabilistic phrase-based model to translate between the (then available) 11 languages of the EPC. As is shown in Figure~\ref{plot:BLEUcor}, Koehn's BLEU scores for pairwise translations correlate with the pairwise ratios of entropies ($r=0.58$, $p<0.0001$).

For example, Finnish (fi) is a high entropy language ($\tilde{H}(T)=8.35$) compared to English (en) ($\tilde{H}(T)=6.32$). Translating from Finnish to English gives a BLEU score of $21.8$, and from English to Finnish $13$. As pointed out above, this is due to the fact that translating into a higher entropy language means having more ``choice'' - or uncertainty - when translating words (or phrases). So the low performance from English to Finnish is predicted by a low English-to-Finnish entropy ratio ($6.32/8.35=0.76$), compared to the Finnish-to-English ratio ($8.35/6.32=1.32$).    

\subsection{Entropy as a normalization factor}
In Section~\ref{motivation}, some examples were given of studies that use information content in a distributional semantics context. Resnik \cite{Resnik1995}, for instance, builds a similarity measure for concepts and words based on the information content, e.g. $-log\, p(w_i)$. Remember that the entropy as defined in Equation~\ref{eq:entropy} is the \textit{average information content}. It is clear from the range of word entropies in Figure~\ref{H_density} that languages can be strongly biased to have words with either high or low information contents on average. For example, words in Finnish are biased to have higher information content on average than words in English. This bias is a problem for the cross-linguistic application of similarity measures. It can be overcome by normalization using the estimated entropy:

\begin{equation}\label{eq:convertIC}
IC_{unbiased}(w_i)=\frac{-log\, p(w_i)}{\tilde{H}(T)}. 
\end{equation}

\section{Conclusions}
The entropy, average information content, uncertainty or ``choice'' is a core property of words. Words, in turn, constitute fundamental building blocks in computational linguistics. Understanding word entropies is therefore a prerequisite for evaluation and improvement of the performance of NLP systems. 

We have here established convergence points for 21 languages, illustrating that average word entropies can be reliable estimated with text sizes of $>70$K. Based on these findings, we estimated entropies across 1360 texts and 1001 languages. Furthermore, we have shown empirically that there is a strong correlation between block and source entropies across these languages. 

Overall, our results help to understand better the performance of multilingual translation systems, and to make measures in distributional semantics cross-linguistically applicable.     

\section{Acknowledgements}
CB is funded by the DFG Center for Advanced Studies \textit{Words, Bones, Genes, Tools}, and the ERC grant EVOLAEMP at the University of Tübingen.

\bibliography{bibtex_all}
\bibliographystyle{abbrv}

\end{document}